\pdfoutput=1

\documentclass[11pt]{article}

\usepackage{acl}
\usepackage{multirow}
\usepackage{times}
\usepackage{latexsym}
\usepackage{graphicx}
\graphicspath{ {./images/} }

\usepackage[T1]{fontenc}

\usepackage[utf8]{inputenc}

\usepackage{microtype}

\title{Tollywood Emotions: Annotation of Valence-Arousal in Telugu Song Lyrics
}

\author{R Guru Ravi Shanker, B Manikanta Gupta, BV Koushik, Vinoo Alluri \\ \\
        International Institute of Information Technology, Hyderabad \\ \\
        \href{mailto:ramaguru.guru@research.iiit.ac.in}{ramaguru.guru@research.iiit.ac.in},\\ \{\href{mailto:battula.manikanta@students.iiit.ac.in}{battula.manikanta},\href{mailto:venkata.koushik@students.iiit.ac.in}{venkata.koushik\}@students.iiit.ac.in},  \href{mailto:vinoo.alluri@iiit.ac.in}{vinoo.alluri@iiit.ac.in}}

\begin{document}
\maketitle 
\begin{abstract}
Emotion recognition from a given music track has heavily relied on acoustic features, social tags, and metadata but is seldom focused on lyrics. There are no datasets of Indian language songs that contain both valence and arousal manual ratings of lyrics. 
We present a new manually annotated dataset of Telugu songs' lyrics collected from Spotify with valence and arousal annotated on a discrete scale. A fairly high inter-annotator agreement was observed for both valence and arousal. Subsequently, we create two music emotion recognition models by using two classification techniques to identify valence, arousal and respective emotion quadrant from lyrics. 
Support vector machine (SVM) with term frequency-inverse document frequency (TF-IDF) features and fine-tuning the pre-trained XLMRoBERTa (XLM-R) model were used for valence, arousal and quadrant classification tasks. Fine-tuned XLMRoBERTa performs better than the SVM by improving macro-averaged F1-scores of 54.69\%, 67.61\%, 34.13\% to 77.90\%, 80.71\% and 58.33\% for valence, arousal and quadrant classifications, respectively, on 10-fold cross-validation.
In addition, we compare our lyrics annotations with Spotify's annotations of valence and energy (same as arousal), which are based on entire music tracks. The implications of our findings are discussed. Finally, we make the dataset publicly available with lyrics, annotations and Spotify IDs.
\end{abstract}
\section{Introduction}
Sentiment analysis deals with identifying affective connotations of text. Typically, humans annotate a corpus, following which a model is trained to classify new samples \cite{pang2002thumbs, apoorva2017bolly}. Most studies involve a categorical approach in which text is classified as either positive or negative in valence or arousal. Valence refers to pleasantness, while arousal refers to energy. 
One of the most widely used emotion models is Russell's circumplex Valence-Arousal (VA) model of affect \cite{russell1980circumplex} and has been specifically used in tasks involving automatic emotion recognition \cite{ccano2017moodylyrics, malheiro2016emotionally_MER}. Furthermore, it has been demonstrated that this model offers to distribute emotions in a 360-degree, four-quadrant VA space. 

Lyrics play an integral role in evoking emotions and contribute to musical enjoyment. However, they are often neglected in tasks involving music emotion recognition \cite{eerola2009prediction, patra2013automatic}. 
There have been lyrics datasets using a dimensional approach containing both valence and arousal values in English \cite{ccano2017moodylyrics, malheiro2016emotionally_MER}. However, the very few that exist in Indian languages have limited annotations to either valence or arousal but not both. For example, BolLy is a Hindi dataset annotated for valence \cite{apoorva2017bolly}, similar datasets exists for Manipuri \cite{devi2020exploiting}, Bengali \cite{nath2020textual}, Telugu \cite{gangula2018resource} while another study has annotated Telugu lyrics for arousal \cite{reddy2018addition}. Also, most lyrics datasets in Indian languages are not publicly available \cite{devi2020exploiting, apoorva2017bolly, patra2015mood}. We need both valence and arousal dimensions to capture the entire gamut of emotions. 

Telugu is a morphologically rich language which makes automatic emotion identification difficult in contrast to languages poor in morphology, such as English. 
In our study, we create a manually annotated lyrics dataset in Telugu, which contains average valence and arousal perceptual ratings. Further, to identify valence, arousal, and quadrant from lyrics, we employ two methods, Support Vector Machine (SVM) \cite{cortes1995support} with TF-IDF \cite{robertson2004understanding} features and fine-tuning pre-trained XLM-RoBERTa (XLM-R) \cite{conneau2019unsupervised} model for emotion recognition on the dataset. They have been extensively used for text classification, including emotion recognition. SVM is a supervised learning algorithm that projects the input to a higher dimensional plane and finds hyperplanes to differentiate two or more classes. XLM-R is a multilingual pre-trained model for many South Asian languages, including Telugu, and Hindi, amongst others. It uses RoBERTa \cite{liu2019roberta} transformer architecture as its base for pre-training, which improves BERT \cite{devlin2018bert} by training on 
longer sequences and more data. It performs very well on downstream natural language processing(NLP) tasks like sentiment analysis and other natural language inference (NLI) tasks. It is also competitive with strong monolingual models on NLI tasks. It also helps to deal with codemix text \cite{ou2020ynu}.

Recently there have been studies \cite{liew2021cultural, lee2021cross} that use Spotify features to find cultural differences. Since lyrics' emotions are mostly congruent with music emotions, we aim to verify congruence between the lyrical perceptual emotion values and Spotify-retrieved emotion values. This has implications in using Spotify features for music emotion recognition, especially in the context of culturally diverse music. 
We release the Telugu lyrics dataset with average valence and arousal values along with Spotify IDs \footnote{https://developer.spotify.com/documentation/web-api/}. 
\section{Related Work}
Before the availability of large text collection and online resources, sentiment analysis was based only on surveys and public opinions \cite{knutson1945japanese}. 
Now, there are annotated corpora for sentiment analysis in English \cite{maas-EtAl:2011:ACL-HLT2011} as well as in other Indian languages such as Telugu \cite{gangula2018resource}, Hindi \cite{shrivastava2020sentiment} amongst others. Mostly the task has been done on short context texts like tweets \cite{agarwal2011sentiment}, and reviews \cite{gangula2018resource} compared to long context text like poems and lyrics. Both categorical and dimensional approaches for emotion recognition are widely accepted for lyrics. MER \cite{malheiro2016emotionally_MER} dataset contains 180 English songs with manually annotated valence and arousal discrete values from -4 to 4. The existing datasets present in the Indian language's lyrics are limited to either valence or arousal categories. 

The task of emotion recognition has evolved from using simple lexicon-based methods \cite{ohana2009sentiment} to using transformer models \cite{agrawal2021transformer}. Numerous studies use NLP techniques like bag-of-words \cite{el2016enhancement}, TF-IDF features \cite{apoorva2017bolly}, word vectors \cite{nath2020textual}, and models like convolutional neural network \cite{nath2020textual}, recurrent neural networks (RNN) \cite{abdillah2020emotion} for sentiment analysis. 
Recently, fine-tuned transformer XLNet model performed better than RNN-based techniques \cite{agrawal2021transformer} for lyrics emotion recognition task because of its ability to capture longer contexts. Hence, we also compare the performance of context-free (SVM with TF-IDF) and context-based (fine-tuning pre-trained XLM-R) on the dataset. 

Spotify features based on music tracks like valence, energy, danceability, and instrumentalness, amongst others, have been popularly used in many studies \cite{surana2020static, lee2021cross} for various tasks such as analyses on music listening habits on streaming platforms and exploring differences in mood perception, respectively.
However, several studies have evidenced culture-specific differences in emotion perception, and to music, \cite{lee2021cross, saarikallio2021emotions}. Although valence and energy features from Spotify rely on the entire music track and not just lyrics alone, owing to the typical congruence between emotions conveyed by lyrics and musical features, one can expect a moderately high correlation between Spotify features and annotated features. Hence we additionally compare manual annotations of our dataset with Spotify's features. 
\section{The Dataset}
\label{section:dataset}
\subsection{Construction of the dataset}
For the construction of the lyrics dataset, we chose Tollywood (Telugu film industry) songs from playlists that were available on Spotify. 
Telugu songs are generally positive, as noted by \cite{reddy2018addition} and hence we initially observed around 70\% of the songs annotated were positive on valence and high on arousal. 
To balance the number of songs in each quadrant, we chose songs from 9 diverse playlists, such as "Happy Vibes Telugu", "Sad Melodies Telugu", "Sleepy Telugu", and "Angry Telugu", amongst others. 

Lyrics were scraped manually in their original script from Spotify, if available. For those not available, alternative websites such as LyricsTape\footnote{https://www.lyricstape.com/}, Lyrics Telugu\footnote{https://lyricstelugu.in/} amongst others, were used for scraping the lyrics. Each track has an associated unique Spotify ID, based on which we removed 19 duplicates. Finally, a total of 481 song lyrics with their Spotify IDs were scraped. 
\begin{figure}[]
\includegraphics[width=0.9\linewidth]{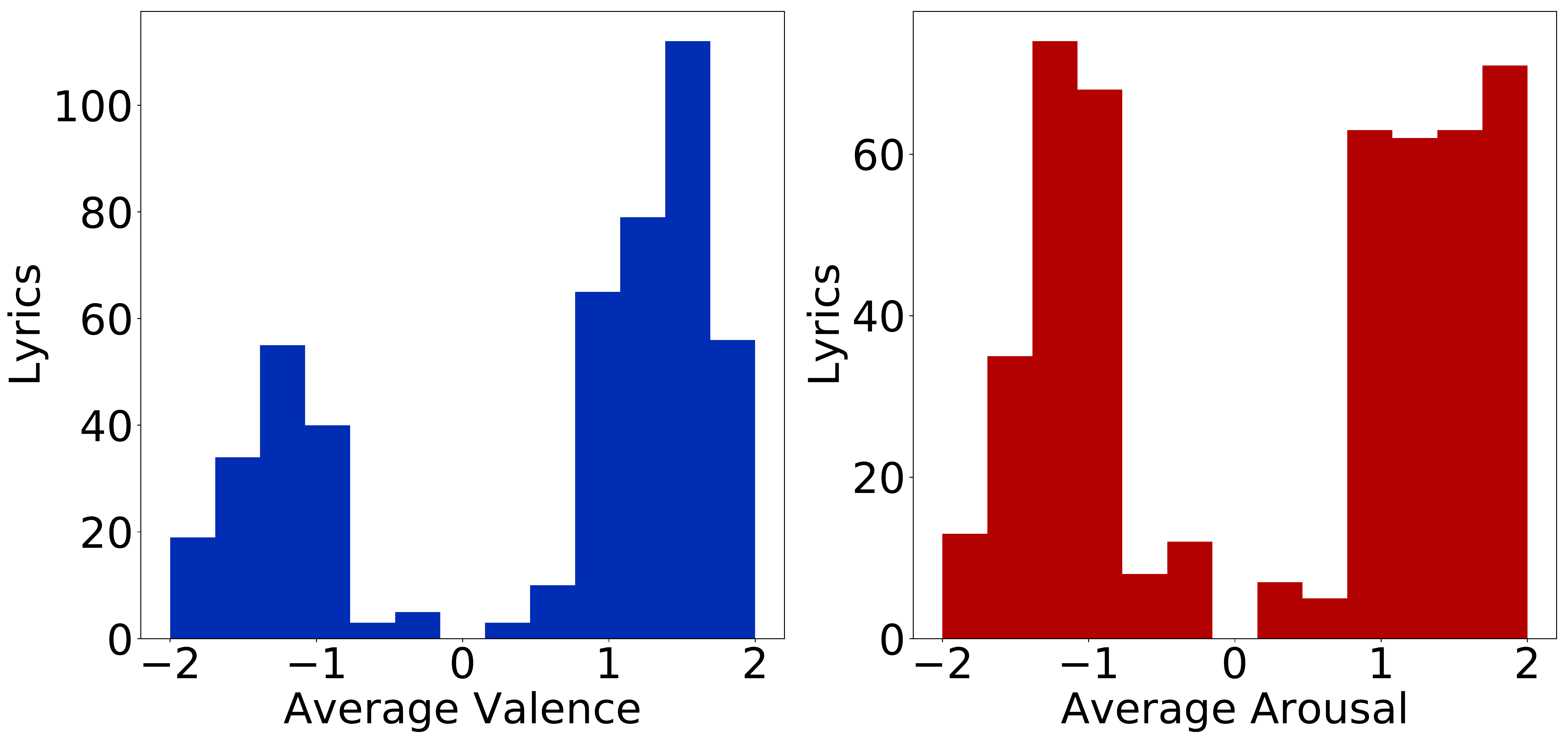}
\centering
\caption{Histogram of Annotated Values}
\label{Histogram}
\end{figure}

\subsection{Annotation}
We use Russell's circumplex model for annotating the lyrics based on valence and arousal. Three annotators, native Telugu speakers, annotated the lyrics of each song for valence and arousal on a discrete five-point scale ranging from -2 to 2. The annotations were solely based on lyrics without listening to the song's audio. 

To check the reliability and internal consistency of the dataset, we calculated Krippendorff’s alpha \cite{krippendorff2011computing} for ordinal data. As a result, acceptable alphas of 0.716 and 0.782 were obtained for valence and arousal, respectively, implying fair agreement among the annotators.

\subsection{Dataset Release Information}
The dataset has been made publicly available at \url{https://tinyurl.com/mu2zkjtc}.
It contains average valence and arousal values for 481 Telugu songs, as shown in Figure \ref{Histogram}. We also provide Spotify ID for each song which helps extract music-related features from Spotify for further analysis. 

\begin{figure}[]
\includegraphics[width=0.9\linewidth]{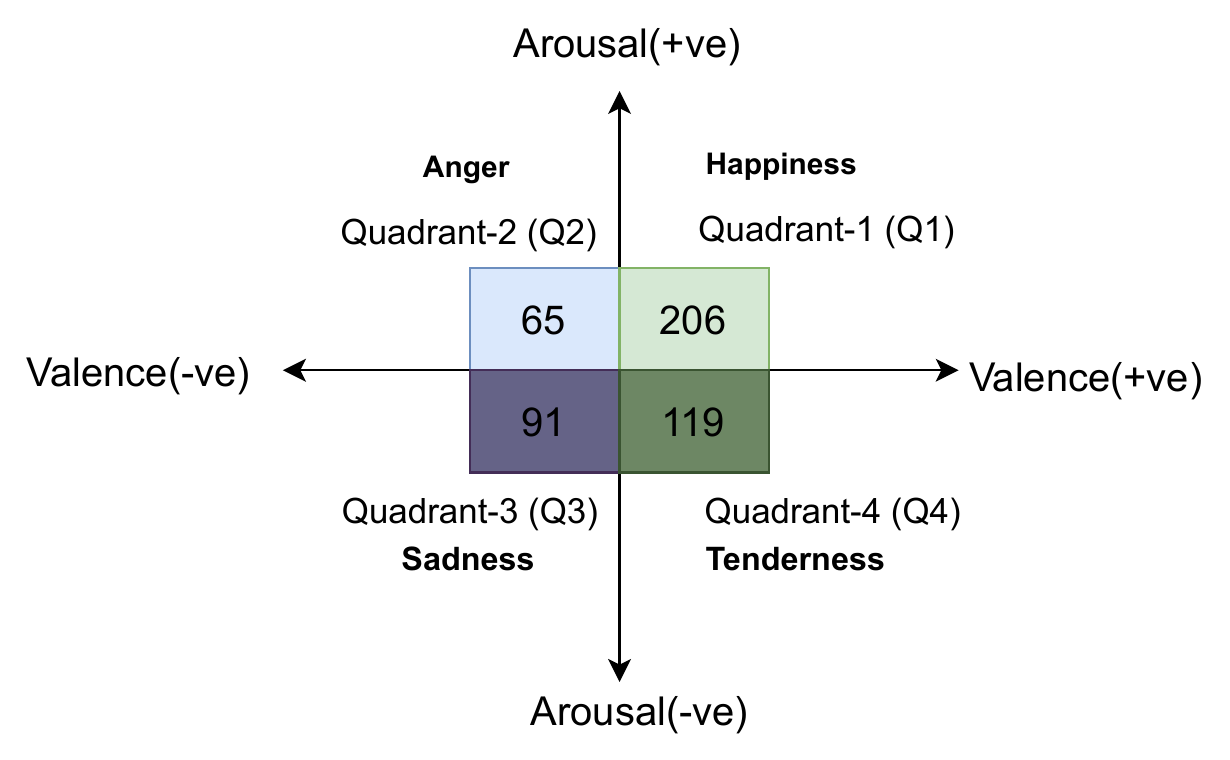}
\centering
\caption{Song distribution in VA plane}
\label{Quadrants}
\end{figure}

Out of the 481 songs, 325 lyrics were perceived as positive valence and 156 as negative valence. Similarly, 271 and 210 lyrics were positive and negative arousal lyrics, respectively. We achieved a final quadrant split, as shown in Figure \ref{Quadrants}. 


\section{Experimentation}
\label{section:experimentation}
\subsection{Methodology}
In order to create automatic music emotion recognition models based on our annotated dataset, we use two supervised machine learning approaches, that is, context-based and context-free classification. To this end, we perform three classification tasks, namely valence classification (VC), arousal classification (AC) and quadrant classification (QC). Though our dataset has average valence and arousal annotations on a continuous scale, we chose to perform classification instead of regression owing to the bimodal distribution of the ratings (See Figure \ref{Histogram}). VC involves the prediction of whether the valence is positive or negative for a given track's lyrics. Similarly, AC predicts whether the arousal is high or low. QC involves predicting the quadrant the lyrics belong to. 

For the SVM-based (context-free) classifications, we use the popularly employed TF-IDF features to represent lyrics to give as input to SVM for the classification task. SVM with linear kernel and TF-IDF have been implemented using ‘scikit-learn’ \cite{pedregosa2011scikit} library with default values. This is an elementary context-free method with basic features of TF-IDF for text classification. 
 
For fine-tuned XLM-R model (context-based), we use the pre-trained(xlm-roberta-base)  for Sequence Classification task from the Hugging face \footnote{https://huggingface.co/docs/transformers/model\_doc/xlm-roberta} library to fine-tune our dataset. We use a learning rate of 2e-6 for VC and AC, and 4e-6 for QC with AdamW optimizer \cite{loshchilov2017decoupled}. We also use the default max sequence length of 512 and a batch size of 8 for our implementation. 
We perform 10-fold cross-validation on our dataset and report the average accuracy (also known as the micro-averaged F1-score) and macro-averaged F1-score (F1).
The macro-averaged F1-score ($\mathcal {F}_{1}$) is given by the formula \ref{F1-score}.
\begin{equation}
    F1_{x} = 2\frac{P_{x}R_{x}}{P_{x} + R_{x}}; \qquad \mathcal{F}_{1} = \frac{1}{n} \sum_{x} F1_{x} 
    \label{F1-score}
\end{equation}
$P_{x}, R_{x}$ and $F1_{x}$ are the standard precision, recall and F1-score of a particular class x, and n is the total number of classes. 

\subsection{Spotify Features}
Spotify provides valence and arousal for each track in addition to several other features like danceability, and instrumentalness, amongst others. Spotify doesn't provide the algorithm used to calculate its features but is based on musical tracks in their entirety. The valence and arousal values from Spotify are continuous and lie in the range of 0 to 1. 
Tracks with a valence or arousal greater than 0.5 are considered to be positive or high, and less than 0.5 are considered to be negative or low, respectively. 
To examine the association between the average annotated VA values, and Spotify's VA values, we performed Spearman's correlation. In order to get quadrant-specific insights, we also perform Spearman's correlation between the same quadrant-wise. 
\begin{figure}[]
\includegraphics[width=0.9\linewidth]{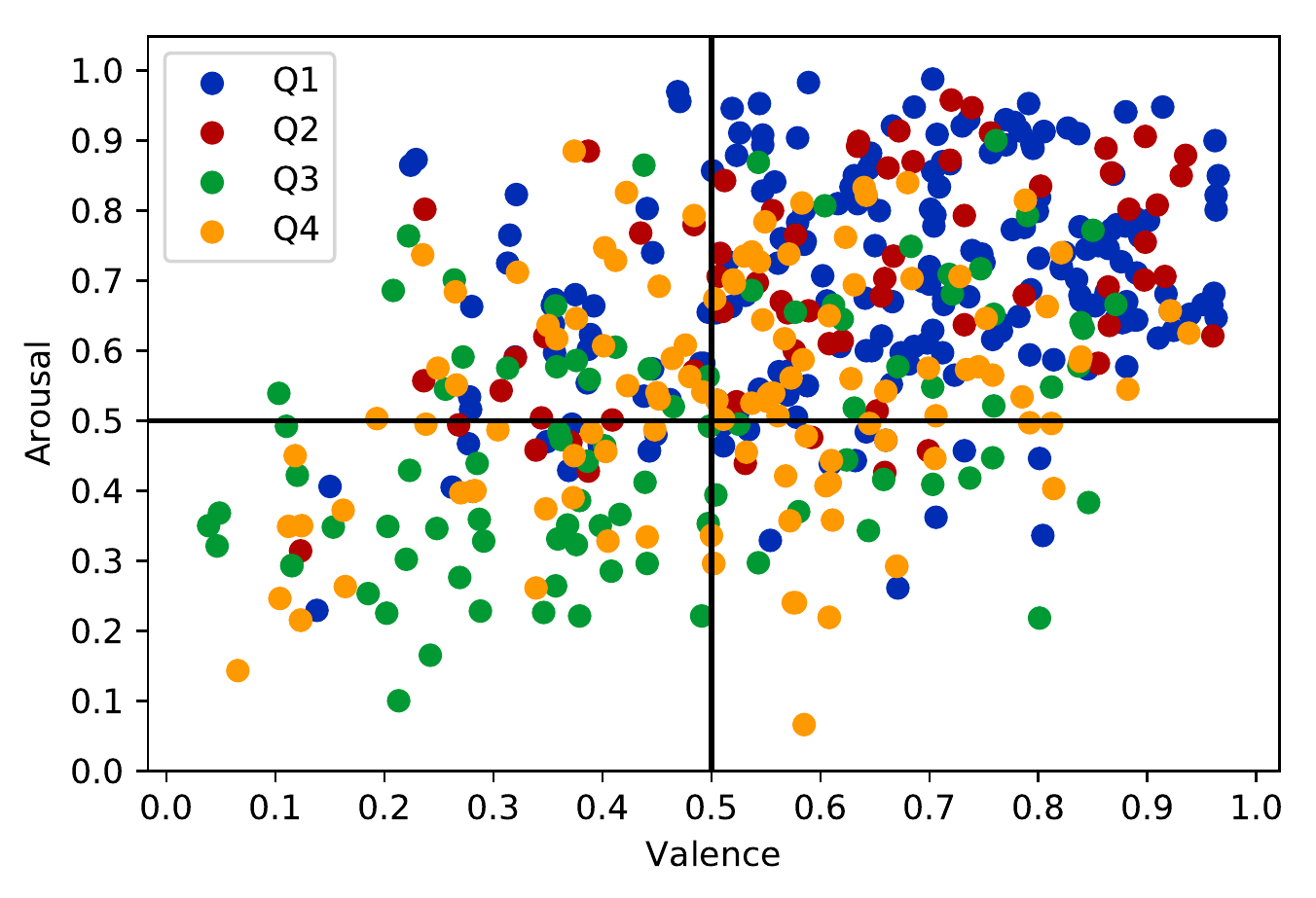}
\centering
\caption{Distribution of Spotify-retrieved VA values by quadrants of annotated VA values}
\label{Scatter}
\end{figure}
\section{Results}
\label{section:results}
\subsection{Classification}
As expected, we observe a far superior performance of the context-based fine-tuning XLM-R model compared to the context-free SVM with TF-IDF, as shown in Table \ref{VC_AC_QC}. 
Nearly ten percentage of song lyrics contain few verses in English or Hindi. This is due to the changing trends of Indian songs in which lyrics are code mixed with other languages. This may explain the better performance of fine-tuned XLM-R \cite{ou2020ynu} compared to SVM with TF-IDF features. 

\begin{table}[]
\centering
\setlength{\tabcolsep}{8pt}
    \caption{Results of valence classification (VC), arousal classification (AC) and quadrant classification (QC) tasks}
\begin{tabular}{|c|c|c|c|}
\hline
\textbf{}\textbf{}           & Method & Accuracy & $\mathcal{F}_{1}$ \\ \hline
\multirow{2}{*}{VC} & SVM             & 69.43\%             & 54.69\%  \\ \cline{2-4} 
                    & XLM-R      & \textbf{80.88}\%           & \textbf{77.90\%}  \\ \hline
\multirow{2}{*}{AC} & SVM             & 68.60\%           & 67.61\%   \\ \cline{2-4} 
                    & XLM-R      & \textbf{81.51\%}           & \textbf{80.71\%}  \\ \hline
\multirow{2}{*}{QC} & SVM             & 48.63\%           & 34.13\%  \\ \cline{2-4} 
                    & XLM-R      & \textbf{62.58\%}           & \textbf{58.33\%}\\ \hline
\end{tabular}
\label{VC_AC_QC}
\end{table}
\subsection{Spotify Analysis}
We observe a small yet significant correlation (r = 0.167, p < 0.01) between annotated and Spotify-retrieved valence and a moderate positive correlation (r = 0.533, p < 0.01) between annotated arousal and Spotify-retrieved arousal. Quadrant-wise correlation analyses revealed significant positive correlations for high arousal quadrants Q1 (r = 0.475, p < 0.01) and Q2 (r = 0.269, p < 0.05) with Spotify-retrieved arousal. For Q4 lyrics, valence annotations correlated (r = 0.225, p < 0.05) positively with Spotify-retrieved valence. 

Though we are comparing lyrics annotations with music annotations, the agreement value is less than expected. From Figure \ref{Scatter}, we can observe the difference in the classification of quadrants based on Spotify-retrieved values and annotated values. 
One possible explanation for low agreement between Spotify and human annotations for Telugu songs is the difference in the mapping of musical features, which could be attributed to cultural differences.

\section{Conclusion and Future Work}
\label{section:conclusion}
In this study, for the first time, we created a dataset of Telugu song lyrics manually annotated for both valence and arousal. In addition, we also provide Spotify IDs. The differences between annotated VA ratings and Spotify VA values highlight the need to build cultural-specific models for better song recommendations, especially since there has been a staggering increase of Indian users on Spotify \cite{ETIndiaSpotify}. Furthermore, this dataset can be used to train models to predict emotions from Telugu text, especially lyrics. Further improvements can be made to balance the dataset across quadrants. This study can be extended to other South Asian language songs for training a multilingual lyrics emotion prediction model. 

\label{sec:bibtex}

\bibliography{anthology,custom}
\bibliographystyle{acl_natbib}




\end{document}